# tSPM+; a high-performance algorithm for mining transitive sequential patterns from clinical data

Authors: Jonas Hügel, M. Sc. [1,2,3], Ulrich Sax, PhD [2,3] Shawn Murphy, MD, PhD [4,5,6] Hossein Estiri, PhD [1,7]

[1] Clinical Augmented Intelligence Group, Department of Medicine, Massachusetts General Hospital, Boston, MA, United States
[2] Department of Medical Informatics, University Medical Center Göttingen, 37075 Göttingen, Germany
[3] Campus Institute Data Science, Georg-August-Universität Göttingen, 37075 Göttingen, Germany
[4] Department of Biomedical Informatics, Harvard Medical School, Boston, MA 02115, United States
[5] Department of Neurology, Massachusetts General Hospital, Boston, MA, United States
[6] Research Information Science and Computing, Mass General Brigham, Somerville, MA 02145, United States
[7] Department of Medicine, Harvard Medical School, Boston, MA, United States

## Abstract

The increasing availability of large clinical datasets collected from patients can enable new avenues for computational characterization of complex diseases using different analytic algorithms. One of the promising new methods for extracting knowledge from large clinical datasets involves temporal pattern mining integrated with machine learning workflows. However, mining these temporal patterns is a computational intensive task and has memory repercussions. Current algorithms, such as the temporal sequence pattern mining (tSPM) algorithm, are already providing promising outcomes, but still leave room for optimization. In this paper, we present the tSPM+ algorithm, a high-performance implementation of the tSPM algorithm, which adds a new dimension by adding the duration to the temporal patterns. We show that the tSPM+ algorithm provides a speed up to factor 980 and a up to 48 fold improvement in memory consumption. Moreover, we present a docker container with an R-package, We also provide vignettes for an easy integration into already existing machine learning workflows and use the mined temporal sequences to identify Post COVID-19 patients and their symptoms according to the WHO definition.

## Introduction

While the primary functionality of Electronic health records (EHRs) is to capture patient data for billing and communication purposes, as research data source, EHRs can provide insights about patient journeys and understanding of complex diseases [1]. Leveraging this information has become feasible by the rapid growth in the availability of computational power and development of new analysis methods. This allows for new methods regarding disease prevention, control, population health management [2], [3], diagnosis of (rare) diseases [4]–[6], treatment options [7]–[11] and drug-development [8], [12] by harnessing big data analytics.

There are a few challenges, such as harmonization and interoperability [13], noisiness [14], [15], availability of computational power, models and data [15], [16] and privacy and security [16], [17], that need to be addressed when working with big data in healthcare. Nevertheless, the large amount of healthcare data presents a valuable resource that, once properly utilized, has the potential to transform patient healthcare, research, and population health [18], [19]. While we have not yet fully tapped into the immense potential of big healthcare data, there are already successful approaches in place, such as machine learning, association rule mining and temporal pattern mining, that are making a significant impact.

This paper presents multiple significant contributions. We introduce an optimized and enhanced implementation of the transitive sequential pattern mining (tSPM) algorithm [20], [21], referred to as tSPM+, for mining transitive sequential patterns from time-stamped clinical data. Estiri et al. [20], [21] introduced an innovative approach for mining transitive (temporal) sequence patterns (tSPM) from electronic health records, which proves beneficial for enhancing signal detection in various machine learning models [20]–[22]. In the year 2021, the tSPM algorithm was recognized as a significant contribution to the field of clinical informatics [23].

This implementation is based on a C++ library wrapped within an R-package, delivering notable improvements in both speed and memory consumption compared to the previous implementation. Specifically, tSPM+ exhibits a speedup up to factor ~920 and ~48-fold reduction in memory consumption. Additionally, the R-package provides a functionality to split the dbmart in chunks with an adaptive size to fit the available memory limitations.

The substantial acceleration of the algorithm unlocks new potential use cases, particularly in leveraging temporal sequences and their durations to simplify complex tasks such as identifying patients with rare or complex diseases, including conditions like Post COVID-19, commonly known as long covid [24]. To demonstrate the application of tSPM+ in such scenarios, we provide a detailed vignette illustrating the implementation of one of these tasks. Specifically, we showcase how to identify patients with Post COVID-19 and associated symptoms within a synthetic database.

Furthermore, we highlight the seamless integration of the tSPM+ algorithm into existing machine learning workflows. By outlining the steps required to incorporate tSPM+ effectively, we offer researchers a straightforward approach to harness the algorithm's capabilities within their established frameworks. To facilitate easy access and reproducibility of our work, we provide a Docker container encompassing an RStudio instance pre-installed with tSPM+, synthetic data, and the accompanying vignettes. This container grants researchers and readers an accessible entry point and ensures easy reproducibility of our findings.

# Background

## Association rule mining

The field of data mining has witnessed significant advancements in extracting knowledge and patterns from extensive databases [25]–[30]. One specific area within data mining is association rule mining (ARM), which aims to extract rules that capture associations, correlations, or frequent patterns among entries in a given database [29], [31]. Since its introduction by Agrawal et al. [29] in 1993, initially for analyzing market basket data, ARM has evolved into an active and extensive research area in data science, encompassing diverse data sources [25]–[28], [30]–[35]. Recently, Shahin et al. [25] conducted a systematic review and identified three commonly employed ARM algorithms: Apriori [29], FP-Growth [36] and Eclat [37]. Over the years, these algorithms have undergone numerous enhancements and adaptations [38]–[43].

Although general association rule mining typically overlooks temporal relationships among individual data entries [44], EHR data inherently possesses temporal dependencies. Consequently, temporal pattern mining techniques are employed to account for such relationships. Sequential pattern mining (SPM) represents a subtype of temporal pattern mining that incorporates the order of entries in the database, including their temporal aspects, while extracting frequent patterns [45]. Within the healthcare domain, SPM serves as a prevalent technique for decision support systems and as input for machine learning algorithms. Leveraging sequential patterns, instead of considering individual entries, facilitates enhanced signal detection for certain machine learning algorithms, making it a widely adopted approach in healthcare [9], [20], [21], [46]–[49]. In some cases, SPM algorithms account for the duration of the sequences. Notably, temporal pattern mining encompasses more than just sequential pattern mining and encompasses extensive subfields such as time series data analysis [50].

While ARM and SPM algorithms offer distinct perspectives on data analysis, they both suffer from shared drawbacks [51]. Their application to larger databases demands substantial computational resources due to their inherent complexity [51]. Moreover, the reliability and accuracy of their outcomes rely heavily on the quality of the input data, making the presence of noise and incomplete data, which are prevalent in medical datasets, particularly influential. Furthermore, the well-established challenge of safeguarding data privacy in the medical domain must be carefully considered when employing ARM and SPM algorithms for medical data analysis. However, overcoming these obstacles can yield valuable insights and enable the exploration of complex research inquiries, ultimately contributing to the enhancement of patient care and well-being [21], [22], [36], [40], [46], [48], [50], [51].

## Transitive sequential pattern mining (tSPM) algorithm

Implemented in the R programming language, the tSPM algorithm operates on patient data structured as a simple table, encompassing the patient number, date, and clinical representations from the database, each denoting the clinical feature space X, hence referred to as 'phenX' in abbreviation. This table adheres to the MLHO [52] format and is referred to as a dbmart.

The tSPM algorithm [20], [21] compasses three key steps. First, it extracts all phenX entries for each patient, sorting them based on their dates to establish a temporal order.
Second, tSPM iterates through the sorted phenX entries and generates sequences that initiate with the current phenX and conclude with another phenX having a later date. This process mines $((n-1)(n))/2$ sequences per patient, where n represents the number of entries for the patient in the dbmart. Given an average of ~400 entries per patient and a cohort of 5000 patients, the tSPM algorithm generates a staggering 399,000,000 sequences.
Consequently, the inclusion of a third optional step becomes highly recommended, involving sparsity screening to mitigate the sequence count. Estiri et al. utilize the Minimize Sparsity Maximize Relevance (MSMR) algorithm [20], which employs a straightforward sparsity screening and employs joint mutual information to discard sparse sequences prevalent in small patient subsets. Fig 1. shows the pseudocode for the tSPM algorithm.
Subsequently, Estiri et al. employ the extracted sequences as input for various machine learning tasks [20]–[22], consistently outperforming alternative approaches. While the combination of tSPM and machine learning tasks yields superior signal detection compared to the conventional approach of using phenX as direct input for machine learning [22], the tSPM algorithm leaves potential for improvement concerning memory consumption and runtime. Furthermore, it is important to note that the tSPM algorithm does not provide information regarding the duration of a sequence, specifically the time difference between the dates of the two phenX entries.

```
Require: dbmart
  sort(dbmart, by(patient_num, date))
  sparseSequences = new List()
  for all patient p ∈ dbmart do
    for all phenx x in p do
      for all phenx y in p with y.date ≥ x.date do
        sparseSequences.add(createSequence(x,y))
      end for
    end for
  end for
  nonSparseSequences = sparsityScreen(sparseSequences)
  return  nonSparseSequences
```

**Figure 1: The pseudocode of the basic tSPM algorithm.**

In the following sections we present tSPM+, an optimized implementation of the tSPM algorithm as a C++ library available as an R package. This yields substantial speed and memory improvements compared to the original version. and allows for more complex uses-cases. These are described in two vignettes, where we highlight a seamless integration in a machine learning workflow as well as a scenario to leverage the mined sequences for Post COVID-19 detection.

For accessibility and reproducibility, we provide a Docker container with tSPM+, synthetic data and the aforementioned vignettes, ensuring easy access and replication.

# Methods

While the original implementation of the tSPM algorithm achieved good results, we recognized the need for a more performant implementation. Optimizing its performance enables us to sequence more patient data allowing for more complex analyses revealing more useful and precise information for downstream phenotype modeling. Additionally, integrating the duration of the sequences adds new dimensions to our analyses and enables even more complex use cases, such as the implementation of the long covid definition.

## transitive Sequential Pattern Mining plus (tSPM+) algorithm

The tSPM+ algorithm follows the same fundamental principles as the tSPM algorithm. It constructs sequences by combining each entry of a patient with all subsequent entries, as outlined in the tSPM section. Notably, the algorithm also captures the duration of these sequences, the time intervals between entry dates, expanding the potential of the generated sequences. Consequently, the data must adhere to the MLHO format to support these functionalities. To optimize memory efficiency, the tSPM algorithm either discards the description column in the preprocessing step or necessitates its removal in before.

To facilitate an efficient implementation, we have developed the tSPM+ algorithm as a high-performance C++ library. This implementation can be directly integrated into low-level C++ programs or encapsulated within higher-level languages such as R or Python. The C++ library encompasses not only the tSPM+ algorithm itself but additional auxiliary functions that have demonstrated utility when working with sequences.

By implementing the tSPM+ algorithm as a C++ library, we capitalize on the advantages of leveraging native data formats and performing faster and more efficient operations compared to higher-level languages. Consequently, we made the decision to store data as a numeric representation, albeit with the trade-off of requiring lookup tables for later translation to their original forms. During the creation of these lookup tables, we assign a running number, starting from 0, to each unique phenX and patient ID. This number is stored as a 32-bit unsigned integer, enabling us to use the patient ID as an index in arrays. Crucially, this numeric representation facilitates the storage of phenX pairs as a distinct hash function that is easily reversible. To construct a sequence, we append leading zeros to the end phenX, resulting in a 7-digit number. We then append this number to the first phenX, creating a unique numeric sequence for each phenX pair. This representation can be effortlessly reverted back to its original form and is interpretable by humans, provided the number of digits for the last phenX is known. Furthermore, it allows us to store the sequence as a 64-bit integer (long long). For a more detailed explanation of the sequence creation process, refer to Figure 2.

The duration of a sequence can be stored in multiple ways. We decided to store the duration of a sequence in days as default, but the unit can be changed via a parameter. Using days allows us to incorporate the duration into the number that represents the sequence. Therefore, we utilize cheap bitshift operations to shift the duration on the last bits of the sequence. Nevertheless, we decided to store the duration in an extra variable to ease the program flow, but leverage this feature in some helper functions, e.g. when calculating duration sparsity. Since the duration is stored in days using unsigned 32 bit integers, it reduces the memory footprint further.

While the numeric representation significantly contributes to the substantial memory reduction, its benefits extend to the use of numerical operators, which allows for fast comparison of the individual values. Nevertheless, the most acceleration arises from the parallelization with OpenMP [53]. The parallelization of the tSPM+ algorithm is straightforward by simultaneously creating the sequences for multiple patients in different threads. This requires sorting the dbmart after patient id as first and date as the second criterion to ensure that each patient is one chunk of entries. For an efficient parallel sorting we leverage the in-place super scalar sample sort (ips4o) algorithm from Axtman et al. [54]. Additionally, the entries in each chunk are chronologically arranged, enabling the creation of all sequences for a phenX by iterating over all subsequent phenX in the same chunk. Consequently, to harness parallelization, we distribute the patient chunks over multiple threads storing the created sequences in thread-specific vectors. This strategic design mitigates resource-intensive cache invalidations, thus optimizing performance.

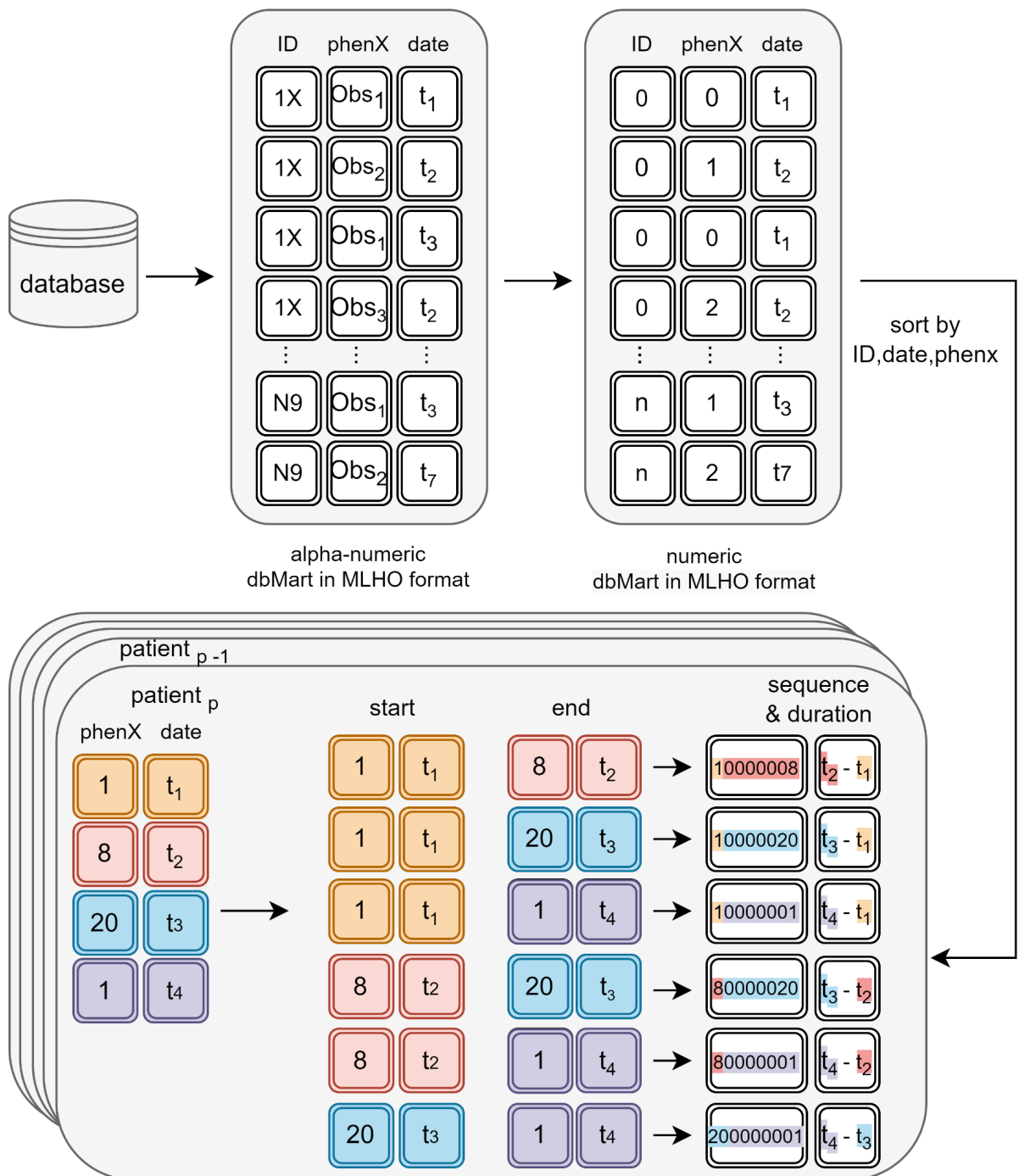

**Figure 2: The workflow to mine the transitive sequences.** At first, the data is extracted from the database and transformed into the MLHO format. After transforming it to numeric, the dbMart gets sorted and the sequences are created for each patient. Each phenX for a patient is coded in a different color. We highlighted the parts (substrings and duration) of the created sequence in the color of the corresponding phenX to visualize how the phenX can be easily extracted from the sequence.

Merging these vectors results in a huge vector of sparse sequences. Sparse sequences occur only for a small number of patients. By removing them, keeping only significant sequences, we preempt overfitting in subsequent machine learning applications. The simplest way to identify sparse sequences is to count the occurrences of a sequence and remove it when the count is less than the threshold. To optimize performance in the parallel processing, we again leverage the ips4o algorithm from Axtman et al. [54] to sort the sequences by their id.

Afterwards, a sophisticated approach is applied to methodically mark sparse sequences before removing them. We first determine the start positions within the vector for each sequence, allowing us to divide it in equal chunks for concurrent processing on multiple parallel threads. In each thread, we iteratively calculate the number of each sequence, by subtracting the start position of the next sequence from the current. If this number is less than the sparsity threshold, we label this sequence for removal by assigning the maximal possible value to the patient number. Once all sequences are labeled, we sort them by their patient id. Subsequently, we determine the first occurrence of the maximal integer value as patient id and erase all values after this entry.

This strategy optimized the number of memory allocations by minimizing its frequency to one. Additionally, the sequence chunks are large enough to mitigate cache invalidations, when altering patients numbers. Finished through shrinking the vector to its new size, we retain only non-sparse sequences, effectively refining the sequences.

## R package:

In order to enhance accessibility to the underlying low-level C++ library, we developed a user friendly R-package. It encapsulates the performant C++ functions making them easily available and usable in the R environment. Rcpp [55] and RcppParallel [56] are widely adopted R packages to interfacing C++ functionalities are often harnessed to speed up and parallelize functionalities in R packages. Consequently, we chose them to facilitate a seamless integration of the tSPM+ C++ library.

Given that tSPM+ exclusively applicable to numeric data, the R package incorporates a utility function to convert alpha-numeric dbmarts to their numeric counterparts and the corresponding look-up tables.

Furthermore, the R-package provides a utility function to enable the adaptive partitioning of the dbmart based on the available memory and the number of created sequences. Applying this approach segregates the data into manageable chunks, which can be sequenced separately. Thereby it enables the sequencing of phenotypes on ressource-restrained platforms, like laptops. This functionality is particularly relevant since the maximum number of entries in an R vector is limited to $2^{31}-1$ entries [57]. This threshold can be swiftly reached when sequencing substantial patient cohorts with multiple tens of thousands patients.

To enhance useability the R-package is accompanied by two instructive vignettes. Each vignette encompasses illustrated code and comprehensive explanations on significant use cases for transitive temporal sequences. These use cases are demonstrable either with provided synthetic example data or data from the linked dependency packages in the vignettes. The first vignette guides the user through integrating the sequencing process into the MLHO machine learning workflow [52], [58]. In contrast, the second vignettes showcased the synergistic utilization of sequences and utility functions to address current challenges, for example, implement complex disease definitions, such as the WHO definition of Post COVID-19.

# Benchmarks

We performed multiple benchmarks to measure the performance of tSPM+. One to compare tSPM+ with tSPM and another one to analyze the possible performance. Since not only the data set characteristics, such as max or average number of phenX per patient and number of patients, might influence the performance of the algorithms, but also the scheduling of the operating system and background processes, we performed 10 iterations of all benchmarks and reported the average, as well as the min/max values for memory consumption and speed. All benchmarks were performed on a machine with Ubuntu 22.04.2 LT, 2 Intel© Xeon© Gold 5220R CPUs @ 2.2GHz, each with 24 Cores and 48 Threads, and 256 GB of available memory. We used R 4.1.2 and compiled the source code using gcc version 11.3.0. We used the time [59] program to measure runtime and maximal memory consumption for each iteration of the tSPM(+) calls.

The benchmark is orchestrated through a bash script, which executes the different R scripts iteratively, for a total of 10 cycles. These scripts encompassed:

1. tSPM without sparsity screening
2. tSPM with sparsity screening
3. tSPM+ in-memory with sparsity screening
4. tSPM+ file-based with sparsity screening
5. tSPM+ in-memory without sparsity screening
6. tSPM+ file-based without sparsity screening.

Within each R script the data was loaded and the corresponding algorithm invocated. The measurement protocol included total runtime and memory consumption as mentioned before, and additionally, the runtime measurement for data loading, sequencing and sparsity screening, if applicable, within the R scripts.

On the one hand, we include the transformation into a numeric representation into the benchmark, because it is a preprocessing step that distinguishes the tSPM and tSPM+ algorithms. On the other hand, we excluded the transformation into the MLHO format from the measurements due to being required by both algorithms.

The bash and R scripts are embedded in the available docker container, as well as in the corresponding GitHub Repo (https://github.com/JonasHuegel/tSPMPlus_benchmarks).

Furthermore, this repository stores a detailed list of the used R packages, their dependencies including the corresponding version numbers Despite the potentially reduced runtime and memory demands of the C++ implementation, we benchmark the R version of tSPM+ algorithm to enhance the comparability with the original tSPM implementation.

## Comparison Benchmark

We analyze the performance of the original tSPM with the tSPM+ algorithm on real world data that were already used together with the old tSPM algorithm in an older AD study [22] to evaluate the performance in a real world setting.

We used the patient data from 4985 patients with an average of 471 entries per patient from the Mass General Brigham Biobank. The Mass General Brigham Institutional Review Board (protocol# 2017P000282) allows the use of the biobank data as per the Biobank Consent signed by all participants in the MGB Biobank.

Following the protocol of the previous study [22], we only kept the first occurrence of a phenX per patient, e.g. when a discarded phenX occurs in the next sequence for a patient, we do not store that sequence. We did this to account for the number of created sequences and the required computational resources of the original tSPM algorithm. Deviating from the previous study, we employed only the sparsity screening from the MSMR function [20] with the tSPM algorithm, but excluded the Joint Mutual Information to select the most relevant features. The tSPM+ library provides a native sparsity function, hence we applied it in the benchmark.

## Performance Benchmark

The second benchmark measures the achievable performance and is performed on the 100k Covid-19 synthetic data set from Synthea™ [60], [61]. After extracting data for ~125 000 synthetic patients and reducing it to 35 000 patients with an average of 318 entries, we stored it in the MLHO format. The reduction of the dataset deemed necessary as the C++ tSPM+ algorithm mined an excessive number of sequences causing failure during the transformation into an R dataframe. This arose from R limiting the number of elements per vector, capped at $2^{31}$-1 elements [57]. While employing adaptive partitioning is a viable approach, we consciously opted against it. Implementing it would introduce extra iterations of the sequencing process without substantial benefits and increasing the runtime linear.

# Results

## Implementations of tSPM+

### The C++ library

The C++ library is implemented in C++17 and is published on GitHub (https://github.com/JonasHuegel/tspm_cpp_backend) under the MIT license. While this implementation is not a direct usable command line tool, it is accompanied by an runnable example file to demonstrate how to include the library in other programs. Moreover, the library encompasses a native function for the sparsity screening and a broad array of additional utility functions allowing fast operations on the sequences. These functions facilitate tasks such as extracting functions with given start phenX, end phenX or specified minimum durations. Another function combines these functions and allows to extract all sequences that end with phenX, which is an end phenX of all sequences with a given start phenX.

The tSPM+ implementation offers two distinct operational modes. The first mode is file based, creating a file storing all generated sequences for each patient. The second mode operates completely in memory, providing the sequences as one comprehensive vector.

### The R package

The R-package is published on GitHub (https://github.com/JonasHuegel/tSPMPlus_R) under the MIT license and encompasses the C++ library as a git submodule. The R package is accompanied by two vignettes and a synthetic dbmart providing examples on how to leverage the outstanding opportunities of the tSPM+ algorithm.

### Integrating tSPM+ in the MLHO Machine Learning workflow

Integrating the mined sequences into existing machine learning workflows is a necessity to leverage the full potential of the sequences. Consequently, the first vignette encompasses instructions to integrate tSPM+ into the MLHO Machine Learning framework. It builds onto the original MLHO vignette [58] and demonstrates how to leverage the sequences for the classification tasks instead of raw EHR entries. In the first step, we load the example data from the MLHO package, converting it to numeric and handing it over to the tSPM+ function call to extract the sequences and perform the sparsity screening. The created, non-sparse sequences are handed over to the MSMR algorithm extracting the 200 most significant sequences. Following the original vignette, we are using MLHO to train the classifier on the remaining relevant sequences. Finally, the vignette demonstrates how the sequences reported as significant for the classification task can be translated back to their descriptions to become fully human readable again.

### Leveraging tSPM+ to identify Post COVID-19 patients

The second vignette encompasses a more complex use case of temporal sequences. We highlight in this vignette how the transitive sequences and their duration can be leveraged to identify which patient has which Post COVID-19 symptom according to the WHO definition. To be considered a Post COVID-19 symptom, a symptom must occur after a covid infection and is at least ongoing for two months, if it can not be excluded by another rationale from the patient. Usually the symptoms appear 3 months after the infection or later, but this is not a mandatory criteria for Post COVID-19 [24].

We utilize a modified version of the synthetic Synthea COVID-19 data set [60], which is included into the R package, as example data. At first, we demonstrate how to transform this alphanumeric dbmart to numeric. Afterwards, we leverage a util function of the tSPM+ library to extract all sequences that end with a phenX that is for at least one patient the end phenX of a sequence starting with covid. From this set we exclude all sequences that did not start with covid. Then we start to exclude candidate sequences on a patient level that either occur only once or where the maximal difference of the duration of the sequences with the same end phenX was less than 2. All remaining sequences are candidates which now need to be excluded by other sequences from a patient. Therefore, we sequence all sequences that end with a candidate phenX and compute pairwise correlations between the sequence and the end phenX duration bucket tupel. If a patient had a sequence with a high correlation, even if it is not casualisation, and the corresponding candidate phenX, we removed the candidate phenX for this patient. After we remove each candidate phenX, for which the patient has at least one other sequence that ends with this candidate phenX and has a high correlation and significance, the remaining candidates are Post COVID-19 symptoms for the corresponding patient. Finally, the vignette demonstrates how to convert the numeric sequences to human readable descriptions.

# Benchmark

## Comparison Benchmark

The tSPM+ algorithm massively outperforms the old tSPM implementation in computation time as well as in memory consumption in the comparison benchmark.

The tSPM+ implementation that just utilizes the files, achieved a speed up by factor ~920; from ~12 900 seconds, a little more than three and a half hours, to ~14 seconds and a memory reduction from ~62.62 GB to ~1.3GB, while the tSPM+ implementation working in memory need ~60 seconds and 43.34 GB of memory, a improvement by factor ~210 in speed and ~1.4 in memory usage respectively. We have to note that for the in-memory approach half of the memory was allocated during the transformation from the C++ data structure into an R-data frame and could be avoided when using it in a C++ program.

The large difference between the file based and in-memory implementation of tSPM+ gets completely equalized when we consider the sparsity screening process. Both implementations require around 25Gb of memory and running in around 1 minute, ~56 and ~64 seconds respectively. Therefore they are clearly outperforming the old tSPM implementation with a runtime of ~19020 seconds and ~205 GB memory consumption providing a speed-up by factor ~297 and an eightfold improvement regarding memory consumption.

**Table 1:** shows the average, min and max values for the memory consumption and runtime for all the implementations during the comparison benchmark. We provide a more detailed enumeration for each run in the appendix.

| Implementation | | | Memory consumption (in GB) | | | Runtime (hh:mm:ss) | | |
|---|---|---|---|---|---|---|---|---|
| **Algo** | **Sparsity Screening** | **In-Memory / File-Base** | **Min** | **Max** | **Average** | **Min** | **Max** | **Average** |
| tSPM | without | In-Memory | 62,27 | 62,82 | 62,62 | 3:30:08 | 3:37:26 | 3:34:09 |
| tSPM | included | In-Memory | 201,09 | 207,60 | 205,23 | 5:10:42 | 5:24:08 | 5:17:27 |
| tSPM+ | without | In-Memory | 43,34 | 43,34 | 43,34 | 00:00:58 | 00:01:11 | 00:01:01 |
| tSPM+ | included | In-Memory | 25,89 | 25,89 | 25,89 | 00:01:01 | 00:01:07 | 00:01:04 |
| tSPM+ | included | File-based | 22,26 | 28,10 | 24,34 | 00:00:52 | 00:00:59 | 00:00:56 |
| tSPM+ | without | File-based | 1,33 | 1,33 | 1,33 | 00:00:13 | 00:00:14 | 00:00:14 |

## Performance Benchmark

When running the performance benchmark with 100k patients with an average of 318 entries, every run failed due to an error in the end when converting the used C++ data structure into an R dataframe. This happens because R has a limit of (2^31)-1 entries per vector and we sequenced 7 195 858 303 (close to 2^33) sparse sequences.

Therefore, we rerun the benchmark with only 35k patients and reported the corresponding runtimes and memory consumption.

As in the comparison benchmark, the file-based tSPM+ algorithm without sparsity is the fastest with an average runtime of ~37 seconds and a memory consumption of ~2 GB, outperforming the in memory approach, which required 109 GB of memory and had a runtime of ~ 214 seconds.

Again this massive lead gets lost, when the sparsity screening is applied. The file-based tSPM algorithm as well as the in memory version with sparsity screening requires an average of ~108 GB of memory. The speed advantage melts down to a difference of ~8 seconds with a runtime of ~288 seconds for the in-memory approach and ~280 seconds for the file-based approach. Table X shows the min, max and average runtime and memory consumption of the performance benchmark. We report the more detailed runtime in the appendix.

**Table 2:** shows the average, min and max values for the memory consumption and runtime for all the implementations during the performance benchmark. We provide a more detailed enumeration for each run in the appendix.

| Implementation | | | Memory consumption (in GB) | | | Runtime (hh:mm:ss) | | |
|---|---|---|---|---|---|---|---|---|
| **Algo** | **Sparsity Screening** | **In-Memory / File-Base** | **Min** | **Max** | **Average** | **Min** | **Max** | **Average** |
| tSPM+ | without | In-Memory | 109,63 | 109,63 | 109,63 | 00:03:10 | 00:04:53 | 00:03:34 |
| tSPM+ | included | In-Memory | 106,61 | 108,16 | 108,01 | 00:04:07 | 00:05:12 | 00:04:48 |
| tSPM+ | included | File-based | 108,17 | 108,20 | 108,18 | 00:03:56 | 00:04:59 | 00:04:40 |
| tSPM+ | without | File-based | 2,01 | 2,19 | 2,12 | 00:00:31 | 00:31:00 | 00:03:40 |

### Performance on End User devices

Additionally, we run the tSPM+ algorithms on some end user devices (laptops or workstations). Even on devices with only 4 to 8 cores and less than 16GB of memory we were able to run the tSPM+ algorithm to sequence more than 1000 patients and ~400 entries per patient in less than 5 minutes.

## Reproducibility and availability of the source code and examples

By integrating the source code from the above-mentioned GitHub repositories into a docker-container, we provide low level access to the tSPM+ algorithms as well ensure reproducibility of our benchmarks. The docker container is based on rocker:rstudio and provides an Rstudio instance, where tSPM+, tSPM, MLHO and all dependencies are already pre installed and ready to use. Furthermore, the docker container is encompassed by both vignettes and their required data. Therefore, it provides two examples demonstrating how to use tSPM+ on synthetic data, and additionally a straightforward approach to deploy tSPM+ and MLHO on their own data. The buildfile and the container is available in the following GitHub repository: https://github.com/JonasHuegel/tSPMPlusDocker. Additionally, we froze the versions of the code and the docker container and provide them online at [62].

## Discussion

In summary, the tSPM+ algorithm significantly outperforms the original tSPM. A fraction of the speedup is achieved by replacing slow string operations and comparisons with faster numeric ones. Consequently, we require 128 bit or 16 byte to store a sequence (8 for the sequence, and 4 for the duration and patient id each. This is significantly smaller than when we use strings (characters) for storing all this information.

To allow an efficient parallelization we added additional sorting steps, which also can be done performant in parallel [54]. After the sorting we can access and modify the data in a linear way avoiding costly cache invalidations and other (scheduling) operations, e.g. memory allocations and copying. This approach is commonly used by other high performance implementations [63]–[65]. A good example for this procedure is the sparsity screening, where we at first sorted the mined sequences by their sequence ID and then just needed to iterate over the sequences and count for how many patients they occur.

According to the developers of the IPS^4o algorithm it is currently not possible to compile their algorithm on windows [54], [66]. Nevertheless, linking it against the RcppParallel library [56], which encompasses the Intel oneAPI Threading Building Blocks library [67], ensures the compilation.

Nevertheless, the tSPM+ algorithm has some limitations. The largest one is that it is currently only working with discrete data. Non-discrete data such as weight can be used, if it gets discretized by creating a new phenX for different ranges. Moreover, since it is only working on numeric data, it requires that the original information is stored in look up tables, which either require memory or have to be written to files. Moreover, tSPM+ requires the transformation to numeric data as a preprocessing step, and the transformation back to human readable sequences, after the sequences were mined and processed in the use cases. While the integration into R provides several advantages, it adds additional overhead, especially when transforming the data from the C++ data structure into an R dataframe, which limits the maximum number of sequences that can be mined per run to $2^{31} - 1$.

The tSPM+ implementation empowers researchers to perform high-throughput sequencing of phenotypes without requiring large scaled servers. By demonstrating that the tSPM+ algorithm performs on end user devices, we enable data scientists and other researchers to develop and test AI/ML pipelines with integrated sequencing on devices with less compute power.

Another advantage of the low resource consumption is that it is possible to sequence large numbers of patients and provide the mined sequences to use them in AI models to examine complex diseases.

For example, limited by computational efficiency Estiri et al. [22] were able to sequence the first occurrence of each phenotype in their Alzheimer's Classification task , tSPM+ would now allow to sequence all phenotypes instead of only the first occurrence of a phenX. Furthermore, tSPM+ provides the duration of these sequences adding a new dimension in the analyses.

In their current review, Xie et al. [68] identify the integration of temporal dimension, especially of entries that might occur multiple times per patient, as a current challenge when using EHR data in deep learning. Using sequences mined with tSPM+ might provide an efficient approach to solve this challenge.

Moreover, as we have shown with our Post COVID-19 vignette, we empower researchers to leverage the usage of transitive sequences to implement complex definitions of diseases without writing complex SQL queries to extract these informations from the databases.However, simplifying complex database queries by utilizing temporal sequences is not a novel approach, already in 2008 Wang et al. [69] worked with temporal sequences trying to avoid complex database queries.

Chaichana et al. [70] analyzed how Post COVID-19 was defined in all the Post COVID-19 studies until the beginning of 2023. According to them there is an urgent need for an easy implementation of a uniform Post COVID-19 definition, since most of the studies were using diverging Post COVID-19 definitions. This might be due to the complexity of definition of Post COVID-19 by exclusion and the challenge of implementing this definition in algorithms. We showed in the vignette that there might be a simple way to fulfill this need. This approach still requires clinical validation, which is why we currently work on a larger multi-site study to evaluate this approach. This approach might also be applicable on other large covid data sets, such as the German NAPKON study [71], [72].

McDermott et al. [16] emphasize the need for reproducible models and implementations for Machine Learning approaches in healthcare. By not only providing example data, but as well as a docker container and two vignettes, we contribute to this need and make our work easily reproducible for others.

Moreover, McDermott et al. [16] stress the danger of applying AI approaches only on inhouse data sets or the "same" public data sets when considering generalization. By providing the vignette on how to integrate tSPM+ with MLHO [52], we enable an easier transfer of the tSPM+ sequencing approach and the MLHO AI models to different data sets. The transfer requires the conversion of the data into the MLHO format. However, by providing the R-package with the synthetic data from Synthea [60], [61], we removed the barrier of having a not shareable data set, allowing others to reproduce most of our results.

# Conclusion

In this work, we presented an efficient, extended high-throughput implementation of the original tSPM algorithm. We provide an R package and a docker-container as low level access to this algorithm and a high-performance C++ library which can be included in different languages.

The massive performance boost of tSPM+ allows for new use-cases, like the aforementioned implementation of the Post Covid definitions. This library enables more researchers to analyze their patient data to solve complex research questions.

By providing two vignettes and a docker container, with relevant use cases and sample data, we reduce the entry barrier for other scientists, especially clinicians, which might not be proficient in programming as data and computer scientists, and just desire an easy to use tool to analyze their EHR data using AI.

Further enhancements to the algorithms, such as the integration of non discrete data, would enable additional dimensions of information, and is worth further investigations.

Additionally, the Post COVID-19 use case requires thorough validation,e. g. by a complete study on its own and would grant urgently required insights in this complex disease.

Finally, tSPM+ adds a new dimension with the sequence durations and is not limited to use only the first occurrence of a clinical record as a phenX for the sequencing. Therefore, it might be worth repeating previous analyses, e.g. regarding Alzheimer's Disease, from older publications to extract more knowledge and get more detailed information about the diseases.

The application of tSPM+ is not limited to Alzheimer's Diseases and covid, but is also applicable to data from other disease trajectories with a temporal component, e,g, cancer and cardiovascular diseases.

# Acknowledgment

J.Hügel's work was partially funded by a fellowship within the IFI programme of the German Academic Exchange Service (DAAD) and by the Federal Ministry of Education and Research (BMBF).
This work is partially funded by the National Institute on Aging (RF1AG074372) and National Institute of Allergy and Infectious Diseases (R01AI165535), the VolkswagenStiftung (ZN3424) and the German Research Foundation (426671079).